%% file: main.tex
\documentclass[letterpaper, 10pt, conference]{ieeeconf}  %
\pdfoutput=1

\IEEEoverridecommandlockouts                              %

\makeatletter
\let\NAT@parse\undefined
\makeatother

\usepackage{cite}
\usepackage{amsmath,amssymb,amsfonts}
\usepackage{graphicx}
\usepackage{textcomp}
\usepackage[dvipsnames]{xcolor}
\usepackage{multirow}
\usepackage{tabularx}
\usepackage{booktabs}
\usepackage{makecell}
\usepackage{xspace}
\usepackage{changepage} 
\usepackage{algorithm}
\usepackage{algpseudocode}
\usepackage{caption}
\usepackage{subcaption}
\usepackage{epigraph}

\let\labelindent\relax
\usepackage[inline]{enumitem}

\newif\ifshowcomments
\showcommentsfalse %
\ifshowcomments
    \newcommand{\igor}[1]{\textcolor{red}{\textbf IV: #1}\xspace}
    \newcommand{\greg}[1]{\textcolor{magenta}{\textbf GS: #1}\xspace}
    \newcommand{\fang}[1]{\textcolor{olive}{\textbf JD: #1}\xspace}
    \newcommand{\matt}[1]{\textcolor{blue}{\textbf MW: #1}\xspace}
\else
    \newcommand{\igor}[1]{}
    \newcommand{\greg}[1]{}
    \newcommand{\fang}[1]{}
    \newcommand{\matt}[1]{}
\fi

\usepackage[square, numbers, sort&compress]{natbib}

\usepackage{flushend}

\makeatletter
\@namedef{ver@lineno.sty}{9999/12/31}
\@namedef{opt@lineno.sty}{}
\makeatother

\usepackage[frozencache=true,cachedir=minted-cache]{minted}
\usepackage[most]{tcolorbox}

\usepackage{etoolbox}
\AtBeginEnvironment{tcolorbox}{\footnotesize}

\definecolor{lightgreen}{rgb}{0.56, 0.93, 0.56}
\definecolor{mygreen}{rgb}{0.2, 0.47, 0.15}
\definecolor{moonstoneblue}{rgb}{0.45, 0.66, 0.76}
\definecolor{mypink}{rgb}{1.0, 0.75, 0.80}
\definecolor{myorange}{rgb}{1.0, 0.65, 0.0}
\definecolor{mykhaki}{rgb}{0.76, 0.69, 0.57}
\definecolor{light_cornflower_blue_2}{rgb}{0.64,0.76,0.96}
\definecolor{light_yellow_2}{rgb}{1.0, 0.9, 0.6}
\definecolor{light_green_2}{rgb}{0.71, 0.84, 0.66}
\definecolor{light_green_1}{rgb}{0.57, 0.76, 0.49}
\definecolor{dark_green_1}{rgb}{0.42, 0.66, 0.31}
\usepackage[bookmarks=true,colorlinks=true,citecolor=blue,linkcolor=blue]{hyperref}

\usepackage{xspace}
\usepackage[textwidth=1.5cm,textsize=scriptsize]{todonotes}

\newcommand{\acronym}{Transcrib3D\xspace}
\newcommand{\OurCaP}{CaP+\acronym}
\newcommand{\VanillaCaP}{CaP\xspace}

\usepackage{float}
\floatstyle{plain}
\newfloat{prompt}{thp}{lop}
\floatname{prompt}{Prompt}

\def\BibTeX{{\rm B\kern-.05em{\sc i\kern-.025em b}\kern-.08em
    T\kern-.1667em\lower.7ex\hbox{E}\kern-.125emX}}

\usepackage[compact]{titlesec}

\title{Transcrib3D: 3D Referring Expression Resolution\\ through Large Language Models}

\begin{document}

\newcommand\blfootnote[1]{%
  \begingroup
  \renewcommand\thefootnote{}\footnote{#1}%
  \addtocounter{footnote}{-1}%
  \endgroup
}

\author{Jiading Fang$^{*,1}$, Xiangshan Tan$^{*,1}$, Shengjie Lin$^{*,1}$, Igor Vasiljevic$^2$, Vitor Guizilini$^2$, \\ Hongyuan Mei$^1$, Rares Ambrus$^2$, Gregory Shakhnarovich$^{1,2}$, Matthew R. Walter$^1$\\
\\[-5pt]
$^1$Toyota Technological Institute at Chicago\quad
$^2$Toyota Research Institute
}

\maketitle
\begin{abstract}
\input{sections/abstract}
\end{abstract}

\def\thefootnote{*}\footnotetext{Denotes equal contribution.}\def\thefootnote{\arabic{footnote}}

\input{sections/intro}

\input{sections/related_works}

\input{sections/methodology}

\input{sections/experiments}
\input{sections/conclusion}

\flushcolsend

{
\bibliographystyle{IEEEtranN}
\bibliography{references}
}
\flushcolsend

\end{document}

%% file: sections/abstract.tex
If robots are to work effectively alongside people, they must be able to interpret natural language references to objects in their 3D environment. Understanding 3D referring expressions is challenging---it requires the ability to both parse the 3D structure of the scene and correctly ground free-form language in the presence of distraction and clutter. 
We introduce \acronym, an approach that brings together 3D detection methods and the emergent reasoning capabilities of large language models (LLMs). \acronym uses \emph{text} as the unifying medium, which allows us to sidestep the need to learn shared representations connecting multi-modal inputs, which would require massive amounts of annotated 3D data.
As a demonstration of its effectiveness, \acronym achieves state-of-the-art results on 3D reference resolution benchmarks, with a great leap in performance from previous multi-modality baselines.
To improve upon zero-shot performance and facilitate local deployment on edge computers and robots, we propose self-correction for fine-tuning that trains smaller models, resulting in performance close to that of large models.
We show that our method enables a real robot to perform pick-and-place tasks given queries that contain challenging referring expressions. Code will be available at \url{https://ripl.github.io/Transcrib3D}.

%% file: sections/intro.tex
\section{Introduction}

Comprehending a natural language expression that mentions an object within a given environment is a routine activity for humans. It often occurs as part of a question (e.g., ``Whose jacket is hanging on the black chair?'') or an instruction (e.g., ``Pass me the smaller yellow mug.'').
This capability is important for embodied agents that work with humans to accomplish tasks. While humans excel in this task with over 90\% accuracy on existing benchmarks~\cite{Achlioptas2020ReferIt3DNL}, contemporary methods %
only achieve mediocre accuracy. They typically rely on supervised learning, e.g., training a Transformer module to obtain contextualized embeddings of various modalities (i.e., text, image, and point-cloud), from which the final prediction is made by a small decoder.
\greg{I wasn't sure: is this "procedure" also part of the transformer in such methods? Or a separate mechanism?}
\fang{the "connection" and ""reasoning" modules are implicitly built in one Transformer, there is no separate mechanism. To me, it is part of the reasons why their performance are not that high because it packs too much into one model considering there are not even enough data to train. I wonder how to best illustrate this.}

\begin{figure}[t]
    \centering
    \includegraphics[width=\linewidth, bb=0 0 695 334]{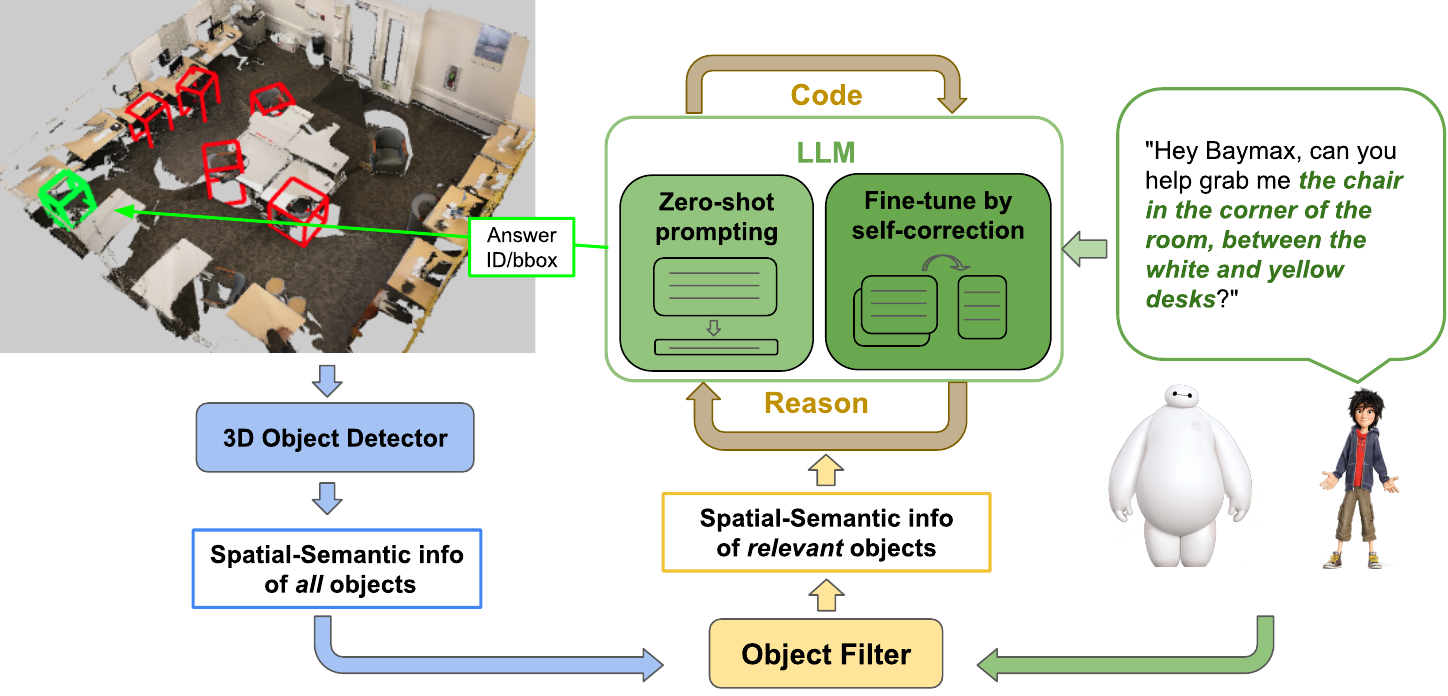}
    \caption{The overall \textbf{\acronym} framework, which takes as input the colored point-cloud and referring expression (in \textcolor{mygreen}{green}), and outputs the ID or bounding box of the referent object. To resolve the referring expression \emph{``the chair in the corner of the room, between the white and yellow desks''}, the framework needs to locate the pillow in the \textcolor{green}{green} box, while all other pillows in \textcolor{red}{red} boxes are distractors. 
    }
    \vspace{-5ex}
    \label{fig:pipeline}
\end{figure}

However, bridging different modalities in a latent space is challenging. For instance, state-of-the-art 2D vision-language understanding models like CLIP~\cite{Radford2021LearningTV} require billions of image-text pairs for training, yet still exhibiting a limited grasp of compositional and relational concepts~\cite{liu2022compositional}. The challenge intensifies in 3D domains, where annotated data is much more scarce~\cite{Deitke2023ObjaverseXLAU}. As a result, the capacity of existing models to perform 3D reference resolution is limited.

On the contrary, philosophers such as Ludwig Wittgenstein argue that our understanding of reality is confined by the language we use, who famously stated, ``The limits of my language mean the limits of my world.''
This concept underpins our approach, wherein we propose to employ \emph{text} as the unifying medium to bridge the gap between 3D scene parsing and referential reasoning.
This approach is grounded in realizing that the challenge of resolving 3D referring expressions can be fundamentally divided into two components: \emph{detection} (identifying objects in the scene) and \emph{reasoning} (associating one of the candidates with the referring expression). 
By harnessing text as a cohesive bridge, we can capitalize on the recent advancements in 3D detection and the enhanced natural language reasoning abilities offered by large language models (LLMs)~\cite{OpenAI2023GPT4TR}.

\begin{figure}[t]
    \centering
    \includegraphics[width=0.9\linewidth, bb=0 0 231 174]{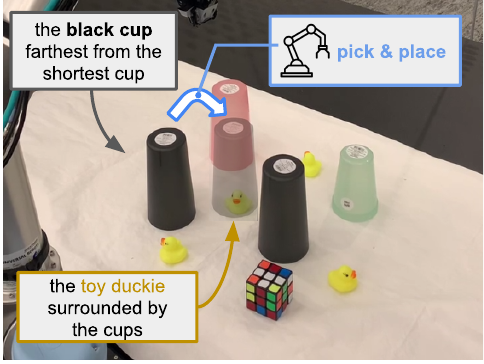}
    
    \caption{\acronym enables a robot to resolve complex 3D referring expressions necessary to to follow pick-and-place instructions. In this example, the robot is tasked with a natural language instruction that includes challenging referring expressions \emph{``cover \textcolor{YellowOrange}{\textbf{the toy duckie surrounded by the cups}} with \textbf{the black cup farthest from the shortest cup}''}.}
    \label{fig:real-robot-demo}
    \vspace{-3ex}
\end{figure}
Specifically, from the results of an off-the-shelf 3D detector~\cite{Schult23ICRA}, \acronym first converts the detected spatial and semantic 3D scene information---the category, location, spatial extent, and color of objects---into texts, thereby creating an object-centric 3D scene \emph{transcript} (hence the name of our method, \textbf{\acronym}). We then filter out non-relevant objects in regard to the query (e.g., information about a trash can is not relevant to the expression ``the white pillow on top of the chair and next to a blue pillow''). Subsequently, we compose a prompt that incorporates the filtered 3D transcript and the referring expression, and process it through an LLM-based reasoning mechanism. The reasoning module incorporates three key elements to make LLMs more effective and generalizable for the task:
\begin{enumerate*}[label=(\arabic*)]
    \item Iterative code generation and reasoning;
    \item Principle-guided zero-shot prompting; and
    \item Fine-tuning from self-reasoned correction.
\end{enumerate*}

We evaluate our method on standard 3D referring expression resolution benchmarks, ReferIt3D~\cite{Achlioptas2020ReferIt3DNL} and ScanRefer~\cite{Chen2019ScanRefer3O}, achieving state-of-the-art performance on both. We also perform real robot experiments (Fig.~\ref{fig:real-robot-demo}) that task a robot manipulator with following natural language commands that require sophisticated 3D spatial-semantic reasoning, demonstrating the practicality of the method.

%% file: sections/related_works.tex
\section{Related Work}

The problem of resolving 3D referring expressions has garnered significant attention of late, in large part due to the introduction of the ReferIt3D~\cite{Achlioptas2020ReferIt3DNL} and ScanRefer~\cite{Chen2019ScanRefer3O} benchmarks. ReferIt3D contains two subsets: SR3D, which consists of template-based utterances, and NR3D, which consists of human-sourced free-form utterances.
Contemporary methods~\cite{Huang2022MultiViewTF,Jain2021BottomUT,Yang2021SAT2S,Chen2022LearningPH} perform 3D referring expression resolution by aggregating different input modalities into contextualized embeddings using Transformer architectures~\cite{Vaswani2017AttentionIA} in an end-to-end fashion.
MVT~\cite{Huang2022MultiViewTF} projects 3D information into 2D to achieve better feature encoding. BUTD-DETR~\cite{Jain2021BottomUT} fine-tunes detected 3D bounding boxes within the Transformer. SAT~\cite{Yang2021SAT2S} uses 2D semantics during training to learn a mapping from the query to its 3D grounding. 
HAM~\cite{Chen2022LearningPH} presents a hierarchical alignment model that learns multi-granularity visual and linguistic representations. Different from these methods, NS3D~\cite{Hsu2023NS3DNG} proposes a neuro-symbolic framework that utilizes a language-to-code model to generate programs, where each module is represented by neural networks. D3Net~\cite{Chen2021D3NetAU} and 3DJCG~\cite{Cai20223DJCGAU} jointly learn 3D captioning and grounding together, where D3Net proposes self-critical training while 3DJCG proposes task-agnostic shared modules and separate task-specific heads. ViL3DRel~\cite{Chen2022LanguageCS} designs a spatial self-attention layer to account for relative distances and orientations between 3D objects. 3D-VisTA~\cite{Zhu20233DVisTAPT} performs pre-training on a dataset of $278$k 3D scene-text pairs, and fine-tunes the model on specific tasks.

In contrast, we propose to connect 3D detection and the LLM reasoning module via a textual representation, which spares us from learning the joint representation of different input modalities from limited 3D annotated data.

\subsection{Grounding Large Language Models}
LLMs trained on Internet-scale text data have shown dominant performance across various NLP tasks~\cite{OpenAI2023GPT4TR}. However, LLMs have to be grounded to answer questions or execute actions in the physical world.
VisProg~\cite{Gupta2022VisualPC} uses in-context learning for LLMs to generate code for 2D image processing tasks.
SayCan~\cite{Ahn2022DoAI} and SayPlan~\cite{Rana2023SayPlanGL}
instead modulate the LLM outputs with a model of the perceived environment (i.e., ``affordances'' for SayCan and scene graphs for SayPlan). 

When it comes to explicitly reason over 3D inputs, one branch of works train multi-modal models that directly incorporate 3D representations into the token library. 3D-LLM~\cite{Hong20233DLLMIT} incorporates distilled 3D features~\cite{Kobayashi2022DecomposingNF} with language tokens. 3D-VisTA~\cite{Zhu20233DVisTAPT} performs multi-modal fusion of language and PointNet++ features~\cite{Qi2017PointNetDH} with self-supervised masked encoding training.

Instead of multi-modal approaches, studies have investigated the use of textual 3D information to enhance downstream tasks. For instance, \citet{Feng2023LayoutGPTCV} employ textual 3D data in conjunction with LLMs to generate 3D indoor scenes. Similarly, \citet{Yu2023CoNavGPTMC} use LLMs for multi-robot navigation, leveraging extracted 3D scene information. We see this body of work as indicative of the promise of textual 3D scene information, and a motivation for our approach. Concurrent with our work, \citet{Yang2023LLMGrounderO3} also use LLMs to reason over textual data, however, their methodology differs in its use of a three-step, task-specific reasoning process as opposed to our general, flexible approach.

\subsection{LLM Reasoning}
There are many techniques proposed to enhance the reasoning capabilities of LLMs~\cite{Wei2022ChainOT, Yao2023TreeOT, Yoneda2023StatlerSL}. The reasoning module of our method is similar in spirit to the general framework of ReAct~\cite{yao2022react}, where each round of code generation is followed by a round of LLM analysis to proceed with the reasoning or debugging, although our context is 3D-specific. Self-correction is also studied in LLM community. In contrast to Reflexion \cite{shinn2024reflexion}, which uses test-time environmental feedback, we instead collect feedback from training samples, and use the corrected samples to fine-tune models.

%% file: sections/methodology.tex
\section{Methodology}

Figure~\ref{fig:pipeline} illustrates our proposed \acronym framework.
Given the input colored point-cloud, \acronym first applies a 3D object detector to generate an exhaustive list of objects in the scene transcribed as text (Section ~\ref{sec:detect_and_transcribe}). The list is then filtered to identify objects relevant to the provided referring expression (Section~\ref{sec:object_filter}). The resulting filtered object information along with the referring expression serve as input to the LLM-based reasoning module for inference. To reach the final answer, the reasoning module includes a ``code interpreter'' mode in which the LLM iterates between code generation and reasoning over outputs from code execution (Section~\ref{sec:code_interpreter}). We consider two options for interfacing with the LLM:
\begin{enumerate*}[label=\arabic*)]
    \item zero-shot with principles-guided prompting, and
    \item fine-tuning from self-reasoned correction
\end{enumerate*}
(Section~\ref{sec:self_correction}).

\subsection{Detect and Transcribe 3D Information}\label{sec:detect_and_transcribe}

Given a colored point-cloud of the scene, we first perform semantic segmentation of objects using Mask3D~\cite{Schult23ICRA}. We associate with each detected object its category based on the semantics, center location, spatial extent according to the 3D bounding box, as well as its mean color. The 3D orientation of an object can also be incorporated using PartNet~\cite{Mo2018PartNetAL}. \acronym compiles the information associated with all detected objects as a list to form an object-centric scene description. The following shows an example of such a scene transcript (exact numeric values and duplicate object categories are replaced by \dots).
\begin{prompt}[h]
\begin{tcolorbox}[
    enhanced,
    attach boxed title to top left={xshift=6mm,yshift=-3mm},
    colback=light_cornflower_blue_2!20,
    colframe=light_cornflower_blue_2,
    colbacktitle=light_cornflower_blue_2,
    title=Spatial Semantic Scene Description for All Objects,
    fonttitle=\bfseries\color{black},
    boxed title style={size=small,colframe=light_cornflower_blue_2,},
]
scene0592: ... Scene center: [...] ... objs list:\\
monitor, id=0, ctr=[...], size=[...], rgb=[...];\\
...\\
box, id=5, ctr=[...], size=[...], rgb=[...];\\
copier, id=6, ctr=[...], size=[...], rgb=[...];\\
floor, id=7, ctr=[...], size=[...], rgb=[...]; \\
wall, id=8, ctr=[...], size=[...], rgb=[...];\\
...\\
armchair, id=15, ctr=[...], size=[...], rgb=[...];\\
...\\
chair, id=19, ctr=[...], size=[...], rgb=[...];\\
...
\end{tcolorbox}
\label{prompt:scene_description}
\vspace{-3ex}
\end{prompt}

\subsection{Pre-Filtering Relevant Objects for Utterance}\label{sec:object_filter}

The aforementioned procedure results in a representation of every object detected in the scene. However, only a small fraction of these objects will typically be relevant to the given referring expression. 
\noindent Given the utterance,
``\texttt{chair in the corner of the room, between white and yello desks}'',
the model identifies the following objects as relevant,
\begin{tcolorbox}[
    enhanced,
    attach boxed title to top left={xshift=6mm,yshift=-3mm},
    colback=light_yellow_2!20,
    colframe=light_yellow_2,
    colbacktitle=light_yellow_2,
    title=Spatial Semantic Scene Description for Relevant Objects,
    fonttitle=\bfseries\color{black},
    boxed title style={size=small,colframe=light_yellow_2,},
]
scene0592: ... Scene center: [...] ... objs list:\\
    wall, id=8, ctr=[...], size=[...], rgb=[...];\\
    ...\\
    armchair, id=15, ctr=[...], size=[...], rgb=[...];\\
    ...\\
    chair, id=19, ctr=[...], size=[...], rgb=[...];\\
    ...
\end{tcolorbox}
\noindent and the remainder as irrelevant.
Simplifying the object list by filtering out irrelevant items (taking into account synonyms and hypernyms) not only reduces processing time and the number of tokens for LLMs, but also facilitates reasoning by reducing content that may potentially be distracting. Such an approach has been shown to improve the efficiency of language grounding~\cite{patki19}.

\subsection{Iterative Code Generation and Reasoning}\label{sec:code_interpreter}

\begin{figure}[!t]
    \centering
    \includegraphics[width=0.8\linewidth, bb=0 0 342 155]{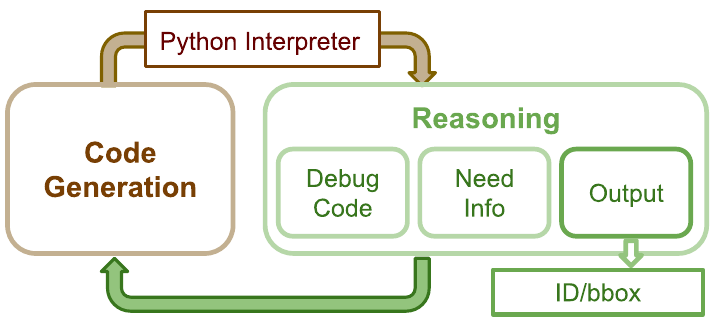}
    \caption{Illustration of the iterative code generation and reasoning process. After the code generation, execution results from a local Python interpreter is fed back to the LLM for further reasoning. The LLM then proceeds to either 1) fix code errors when encountered, 2) generate additional code to obtain more information, or 3) output the referred objects if it has all the needed information. This process continues until the LLM believes the reasoning to be complete.}
    \label{fig:code-gen}
    \vspace{-3ex}
\end{figure}
Compositional reasoning that involves arithmetic calculations, which is crucial for spatial reasoning, is well known to be a weakness of Transformers and LLMs~\cite{Dziri2023FaithAF}. However, most contemporary approaches to 3D referring expression resolution rely on a single forward pass of a Transformer to reach the final answer, which can be limiting in terms of reasoning power. In order to avoid this weakness, \acronym equips the LLM with a Python interpreter and directs the LLM to generate code whenever quantitative evaluations are necessary. \acronym then locally executes the generated code using a Python interpreter and appends the output to the original conversation. The resulting prompt is sent back to the LLM to generate the next response. If there is any error in executing the generated code, \acronym also feeds information about the error back to the LLM for re-generation. This process continues until the LLM believes that the reasoning is complete. Figure~\ref{fig:code-gen} provides an illustration of this process.

We want to emphasize that our design does not distinguish between code generation and chat reasoning steps. Such a framework simplifies the design, allows for flexible reasoning, and makes it easy for fine-tuning. But in turn, it requires the LLM to have the ability to do both, so language models that are specifically trained for only coding or only chatting will not perform the best within our framework.

\subsection{Principles-Guided Zero-Shot Prompting}\label{sec:prompt_principles}

Spatial reasoning can require complex compositional logic that is challenging for LLMs~\cite{Dziri2023FaithAF}. For example, when handling left/right spatial relations, which should factor in the observer's viewpoint, LLMs often naively ground them with smaller/larger values in the $x$-coordinates. In order to overcome these deficiencies, \acronym employs a few general principles to guide LLM reasoning in a zero-shot fashion, including how to 1) use HSL space to match colors; 2) perform vector operations to resolve directional relations; 3) calculate the point-to-plane distance to determine who's closer to a wall, to name just a few.
We find that this set of guiding principles works across 3D referring benchmarks. 
The full prompt will be included in the released code.

\subsection{Fine-tuning from Self-Reasoned Correction}\label{sec:self_correction}

\begin{prompt}[!t]
\begin{tcolorbox}[
    enhanced,
    attach boxed title to top left={xshift=6mm,yshift=-3mm},
    colback=dark_green_1!20,
    colframe=dark_green_1,
    colbacktitle=dark_green_1,
    title=Self-Correction for fine-tuning,
    fonttitle=\bfseries\color{black},
    boxed title style={size=small,colframe=dark_green_1},
]
\colorbox{light_green_1}{Assistant:}\\
Based on the calculated metrics, we can see that chair 49 has the smallest distance to the nearest two walls, which means it is most likely to be in the corner of the room. It also has the smallest distance to the nearest white desk and a relatively small distance to the nearest yellow desk, which means it is likely to be between a white desk and a yellow desk. Therefore, chair 49 is the most likely to be the chair described in the description. Now the answer is complete -- {'ID':49}

\colorbox{light_green_1}{User:}
The correct answer is object 18. Can you double check the information of object 18 and the given prompt and see where you got wrong?

\colorbox{light_green_1}{Assistant:}
Let's revisit the information provided for object 18 and the constraints given in the prompt to identify where the mistake might have occurred.

Object 18 is described as follows:
- Type: Chair
- Center: [-2.98, -3.31, 0.39]
- Size: [0.53, 0.61, 0.81]
- RGB: [60, 58, 50]

The constraints from the prompt are:
1. The chair is in the corner of the room.
2. The chair is between a white and a yellow desk.

Upon reviewing the code execution results, it was noted that chair 18 has a relatively small distance to the nearest two walls, indicating it could be in the corner of the room. Additionally, chair 18 has a very small distance to the nearest white desk (0.66) and a reasonable distance to the nearest yellow desk (2.58), which aligns with the constraint of being between a white and a yellow desk.

\colorbox{light_green_2}{The mistake in the initial analysis was not properly considering} \colorbox{light_green_2}{the combination of both constraints together.} While chair 49 had the smallest distance to the nearest two walls and a very small distance to the nearest white desk, it did not optimally satisfy the condition of being between a white and a yellow desk as well as chair 18 did. Chair 18's proximity to both a white and a yellow desk, along with its corner location, makes it the correct answer according to the given constraints.

\end{tcolorbox}
\caption{An example of self-reasoned correction for fine-tuning. During self-correction, the LLM realizes its mistake during the initial analysis.}\label{prompt:self_correction}
\end{prompt}

Rule-based systems~\cite{Winograd1971ProceduresAA} are effective for relatively simple domains that involve a limited set of spatial-semantic concepts and structured language, but struggle to generalize, particularly to open-world domains~\cite{tellex2020robots}. %
Motivated by the now well known benefits of data-driven alternatives to rule-based methods, we adopt a novel fine-tuning method for LLMs that enables learning beyond the given set of rules (i.e., the general guiding principles) by enabling the model to learn from its own mistakes. In effect, we seek to endow LLMs with introspection capabilities. We do so via the following procedure:
\begin{enumerate}%
    \item Use the prompt with general principles on the training set for the LLM to generate an initial set of answers with elaborated reasoning process.
    \item For any incorrect answer, augment the original prompt with the correct object ID and request the LLM to reflect on why the original answer is incorrect (``What went wrong?''). This is followed by a request to output ``clean'' reasoning for the correct answer.
    \item Gather the reasoning processes of the correct examples, and the re-generated ones of the initially incorrect examples to produce the dataset for LLM fine-tuning.
\end{enumerate}
Prompt~\ref{prompt:self_correction} shows an example of the self-reasoned correction step. Note that after fine-tuning, we no longer include general principles in the prompt. In this way, the system not only incorporates the guiding principles, but further improves itself by learning from self-reasoned correction.

%% file: sections/experiments.tex
\section{Experiments}

We evaluate the effectiveness of \acronym using the ReferIt3D~\cite{Achlioptas2020ReferIt3DNL} and ScanRefer~\cite{Chen2019ScanRefer3O} benchmarks. ReferIt3D formulates 3D referring expression understanding as the multiple-choice problem: given a set of segmented objects in a 3D scene along with a corresponding referring expression, identify the unique referent object from the set, typically containing several instances of the same fine-grained category. 
There are five different types of relations in SR3D, namely ``horizontal'', ``vertical'', ``support'', ``between'' and ``allocentric'', which make up approximately $81\%$, $4\%$, $2\%$, $8\%$, and $5\%$ of the data, respectively. ReferIt3D measures performance in terms of accuracy.  Both Sr3D and Nr3D are
split by “Easy”/“Hard” and “ViewDep”/“ViewIndep”. "Easy" samples has one or none distractors in a scene, while "Hard" samples have two or more. 
The view-dependent samples contain language descriptions
that rely on viewing directions. Unlike ReferIt3D, ScanRefer does not provide object segmentation information, and instead tasks methods with returning the 3D bounding box of the target object, given the query utterance and a colored point-cloud. ScanRefer measures performance in terms of accuracy conditioned on the intersection-over-union (IoU) of the ground-truth 3D bounding box over the predicted one, with thresholds at $25\%$ (acc@0.25IoU) or $50\%$ (acc@0.5IoU).

\subsection{Grounding Accuracy on ReferIt3D}
\begin{figure}[!t]
    \centering
    \begin{subfigure}[b]{0.48\linewidth}
         \centering
         \includegraphics[width=\textwidth, bb=0 0 410 226]{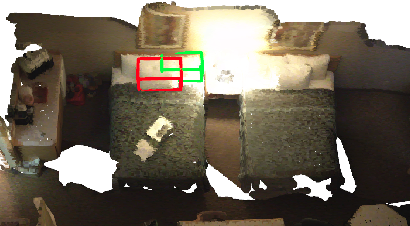}
         \caption{Utterance: ``On the bed with the laptop its the pillow by the light.'' This query requires understanding three constraints together: ``On the bed'', ``with the laptop'', and ``by the light''.} \label{fig:comp_example_1}
     \end{subfigure}
     \hfill
     \begin{subfigure}[b]{0.48\linewidth}
         \centering
         \includegraphics[width=\textwidth, bb=0 0 1456 1127]{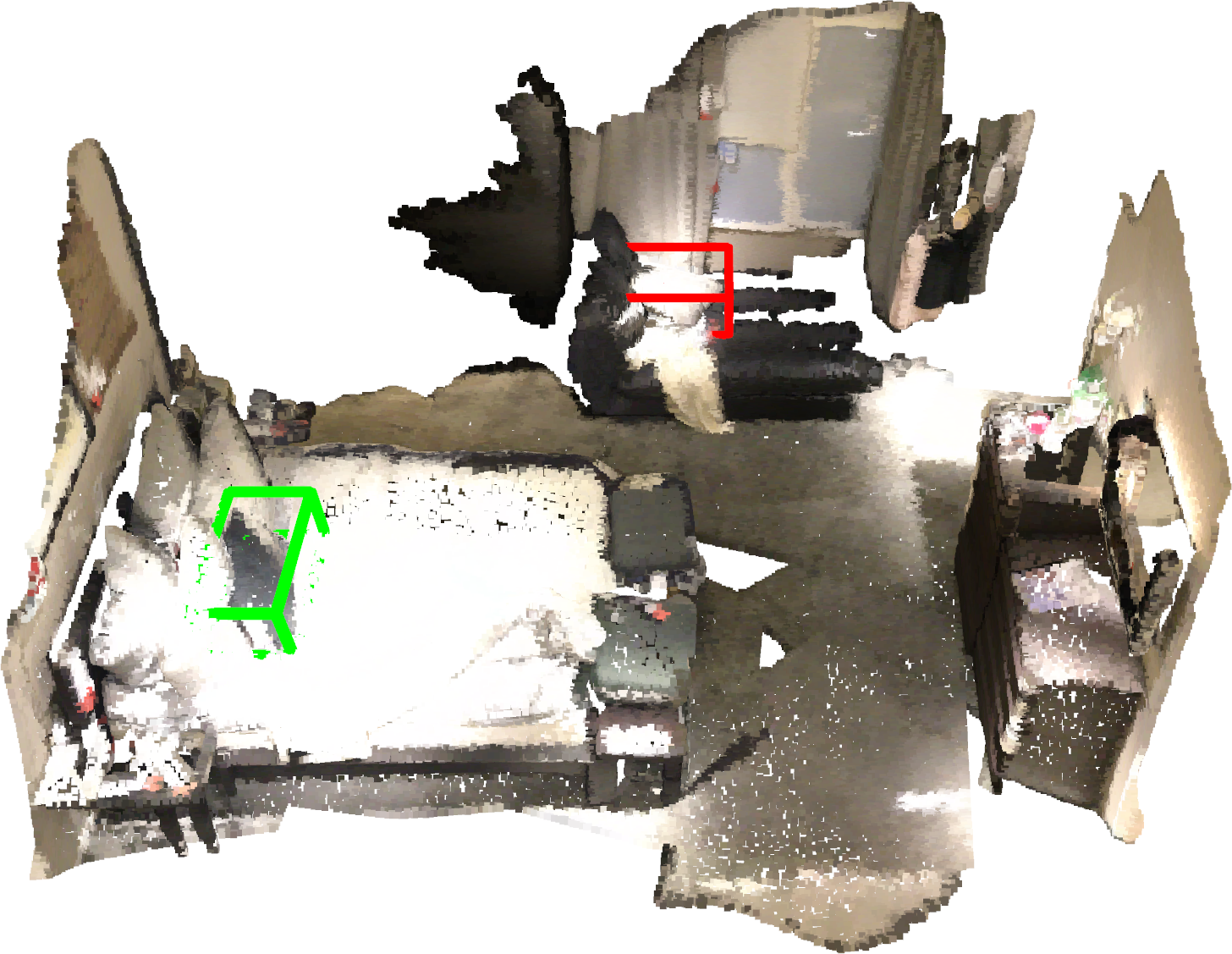}
         \caption{Utterance: ``The foremost pillow on the bed of the group of pillows.'' This query requires understanding that the pillow needs to be among a group.}
         \label{fig:comp_example_2}
     \end{subfigure}
    \caption{Qualitative comparisons between Transcrib3D (ours, in \textcolor{green}{green}) and 3D-VisTA (in \textcolor{red}{red}) on the NR3D dataset.}
    \label{fig:qual_comp_nr3d}
\end{figure}

We test different variations of our method against contemporary baselines on the SR3D and NR3D subsets of ReferIt3D. On NR3D, we evaluate our best model, \acronym (GPT4-P), on the full test set, while other variants of \acronym are evaluated on a subset of $500$ randomly sampled data points. On SR3D, we evaluate all \acronym models on the same subset of $500$ random samples from the test set.\footnote{We use a subset of the test set due to the cost of evaluating all variations of \acronym on the full test set, however we believe that the results would be similar on the full test set due to the templated nature of the utterances.}
Table~\ref{tab:ReferIt3D} presents the results. Figure~\ref{fig:qual_comp_nr3d} also provides a qualitative comparison between 3D-VisTA~\cite{Zhu20233DVisTAPT} and \acronym that highlights the strength of our method with regards to complex reasoning.
\begin{table*}[!ht]
    \centering
    \small
    \begin{tabularx}{\linewidth}{lcccccccccc}
        \toprule
        & \multicolumn{5}{c}{Nr3D} & \multicolumn{5}{c}{Sr3D} \\
        \cmidrule(lr){2-6}\cmidrule(lr){7-11}
        Method & Overall & Easy & Hard & \makecell{View Dep.} & \makecell{View Ind.} & Overall & Easy & Hard & \makecell{View Dep.} & \makecell{View Ind.} \\
        \midrule
        SAT$^\dagger$~\cite{Yang2021SAT2S} & 49.2 & 56.3 & 42.4 & 46.9 & 50.4 & 57.9 & 61.2 & 50.0 & 49.2 & 58.3 \\
        BUTD-DETR$^\dagger$~\cite{Jain2021BottomUT} & 54.6 & 60.7 & 48.4 & 46.0 & 78.0 & 67.0 & 68.6 & 63.2 & 57.0 & 67.7 \\
        MVT$^\dagger$~\cite{Huang2022MultiViewTF} & 59.5 & 67.4 & 52.7 & 59.1 & 60.3 & 64.5 & 66.9 & 58.8 & 58.4 & 64.7 \\
        \midrule
        ViL3DRel$^\mathsection$~\cite{Chen2022LanguageCS} & 64.4 & 70.2 & 57.4 & \textbf{62.0} & 64.5 & 72.8 & 74.9 & 67.9 & 63.8 & 73.2 \\ 
        3D-VisTA$^\mathsection$~\cite{Zhu20233DVisTAPT} & 64.2 & 72.1 & 56.7 & 61.5 & 65.1 & 76.4 & 78.8 & 71.3 & 58.9 & 77.3 \\
        \midrule
        \acronym \scriptsize{(GPT-3.5-NP)} & 33.8 & 42.6 & 25.0 & 26.9 & 36.0 & 79.3 & 82.8 & 70.7 & 70.6 & 80.5 \\
        \acronym \scriptsize{(GPT-3.5-P)} & 46.6 & 56.0 & 37.1 & 29.9 & 51.9 & 80.0 & 80.8 & 78.1 & 82.4 & 79.7 \\
        \acronym \scriptsize{(GPT-4-NP)} & 64.5 & 71.8 & 57.1 & 49.4 & 71.3 & 97.4 & 98.4 & 94.7 & 87.3 & \textbf{98.7} \\
        \acronym \scriptsize{(GPT-4-P)} & \textbf{70.2} & \textbf{79.7} & \textbf{60.3} & 60.1 & \textbf{75.4} & \textbf{98.4} & \textbf{99.2} & \textbf{96.2} & \textbf{98.2} & 98.4 \\
        \bottomrule
    \end{tabularx}
    \caption{Grounding accuracy (\%) on Nr3D and Sr3D. $^\dagger$denotes results from the official benchmarks while $^\mathsection$denotes results reported in the respective papers. ``P'': ``with principles'', ``NP'': ``no principles''. All our models are equipped with interactive code generation. \acronym with GPT-4 and general principles surpasses all baselines by a large margin. 
    } 
    \label{tab:ReferIt3D}
\end{table*}

\subsection{Grounding Accuracy on ScanRefer}
We test the best variant (GPT4-P) of our method with different detection modules against baselines on the ScanRefer benchmark, which unlike ReferIt3D, does not provide methods with the ground-truth bounding boxes during inference. %
In addition to comparing to baseline results reported in their respective papers, we also re-run the checkpoints from 3D-VisTA~\cite{Zhu20233DVisTAPT}, the best baseline method, on the same $500$ random samples from the ScanRefer validation set. Table~\ref{tab:scanrefer} presents the results. We achieve state-of-the-art performance on ScanRefer with both detected or ground-truth bounding boxes. It is worth noting that the performance gain over the baseline is greater with ground-truth compared to detected bounding boxes. This is partly due to the fact that the baseline method is trained with the lower-quality detected bounding boxes, which ironically leads to the ground-truth ones being out of distribution. 

\begin{table}[!ht]
    \centering
    \small
    \begin{tabularx}{\linewidth}{lcccc}
        \toprule
        \multirow{2}{*}{Method} & \multirow{2}{*}{Data} & \multirow{2}{*}{Det.} & \multicolumn{2}{c}{Overall} \\
                                 & & & acc@0.25 & acc@0.5 \\
        \midrule
        ViL3DRel~\cite{Chen2022LanguageCS} & Full & PG & 47.9 & 37.7 \\
        3D-VisTA~\cite{Zhu20233DVisTAPT} & Full & M3D & 50.6 & 45.8 \\
        3D-VisTA~\scriptsize{(re-run)} & Full & M3D & 50.7 & 45.9 \\
        \midrule
        3D-VisTA~\scriptsize{(re-run)} & Part. & M3D & 50.6 & 44.6 \\
        \acronym & Part. & M3D & 51.2 & 44.4 \\
        \acronym + Cam & Part. & M3D & 51.3 & 45.5 \\
        \midrule
        3D-VisTA~\scriptsize{(re-run)} & Part. & GT & 55.6 & 55.6 \\
        \acronym & Part. & GT & 62.0 & 62.0 \\
        \acronym + Cam & Part. & GT & 64.2 & 64.2 \\
        \bottomrule
    \end{tabularx}
    \caption{Grounding accuracy (\%) on ScanRefer. ``Full'': the full validation set of ScanRefer consisting of 5410 samples; ``Part.'': a subset of 500 random samples from the validation set, which is the same for all methods; ``Det.'': the 3D object detection module used in the model;
    ``PG'' stands for  PointGroup~\cite{jiang2020pointgroup}, while ``M3D'' stands for Mask3D~\cite{Schult23ICRA} (where the detection accuracy is 56.7 for PG
    and 73.7 for M3D on the ScanNet dataset for the mAP@0.5 metric~\cite{Schult23ICRA}), and ``GT'' for ground-truth bounding boxes. We test our method and re-run 3D-VisTA on the same subset of 500 samples with the Mask3D detector, GT bounding boxes, and an additional ``+ Cam'' setting, where camera view information provided by the ScanRefer dataset is also included in the scene transcript. Note that ScanRefer allows the use of all provided data modalities and ranks methods on the same benchmark regardless. The zero-shot nature of our method allows ease use of this extra information.} \label{tab:scanrefer}
\end{table}

\subsection{Effects of Fine-tuning Methods}
We study the effects of our proposed approach that involves fine-tuning from self-reasoned corrections. Following Section~\ref{sec:self_correction}, we run \acronym with GPT-4 on 500 samples from the NR3D training set and collect both correct and incorrect examples. For the incorrect examples, we let the LLM re-generate the self-corrected reasoning. We use the combination of correct and self-corrected examples to fine-tune the smaller model (i.e., \texttt{gpt-3.5-turbo}) for 3 epochs. We then evaluate all models on the same subset of 500 random samples from the NR3D test set. Notably, we remove all rule-based prompts from the fine-tuning data so that the fine-tuned models are not constrained to human-designed rules, but are instead adaptive to new rules learned via self-correction.

Note that inference is typically faster with a smaller model. %
A fine-tuned small model with comparable performance to that of a large model on this task would be more desirable for local deployment on edge computers or robots, with additional privacy benefits.
Table~\ref{tab:fine-tuning} presents the results.

\begin{table}[h]
    \centering
    \small
    \begin{tabularx}{\linewidth}{Xccc}
        \toprule
        Models & Total & Easy & Hard \\
        \midrule
        GPT-3.5 (zero-shot) & 33.8 & 42.6 & 24.0 \\
        GPT-3.5 (correct-only fine-tuning) & 60.7 & 52.8 & 41.9  \\
        GPT-3.5 (self-correct fine-tuning) & 61.5 & 54.6 & 80.0  \\
        GPT-4 (zero-shot) & 69.4 & 78.7 & 60.0 \\
        \bottomrule
    \end{tabularx}
    \caption{Performance of fine-tuning models on NR3D. Fine-tuned GPT-3.5 models demonstrate a significant improvement in performance when compared to zero-shot models, closely approaching the capabilities of GPT-4. The model fine-tuned on self-corrected examples sees an increase in performance compared to that fine-tuned on only correct examples, particularly for hard queries. Notably, the fine-tuned models are not provided with rule-based prompts during both fine-tuning and inference time, suggesting that implicit decision rules are learned from examples.}
    \label{tab:fine-tuning}
    \vspace{-3ex}
\end{table}

\subsection{Referring Expressions for Robot Manipulation}

We demonstrate how \acronym supports a robot's ability to follow natural-language instructions for pick-and-place manipulation, particularly when complicated referring expressions are involved. %
As a core capability of robot manipulation, language-guided pick-and-place involves (i) breaking down the language instruction into two referring expressions, one each for the ``pick'' and ``place'' identities; (ii) resolving the referring expressions in the context of the robot's surrounding environment; and (iii) performing the pick-and-place actions. We conduct the pick-and-place task with a Universal Robots UR5 arm equipped with a Robotiq 3-Finger Adaptive Robot Gripper placed in a table-top setting. Figure~\ref{fig:real-robot-demo} visualizes the execution of one utterance using our method. 

To parse the given instruction, we employ a language model-generated program (LMP) from few-shot prompting as in Code-as-Policies~\cite{Liang2022CodeAP}, which is instructed to call the \mintinline{python}{put_first_on_second(arg1, arg2)} function with desired arguments \mintinline{python}{arg1} and \mintinline{python}{arg2}. This approach allows free-form texts as input, which is more flexible and natural than template parsing. The following is an example that is used in the prompt for the LMP:
\begin{minted}[fontsize=\footnotesize]{python}
'# query: Pick up the orange between the apples \
and place it in the bowl with a banana in it.
put_first_on_second("orange between the apples", \
"bowl with a banana in it")'
\end{minted}

The \mintinline{python}{put_first_on_second(arg1, arg2)} function first composes an exhaustive list of objects in the environment along with their spatial-semantic details. To do so, the function employs MDETR~\cite{kamath2021mdetr}, an open-vocabulary object segmentation method, that takes as input an RGB image and an object category, and outputs the 2D spatial attributes of all objects in the scene of that category. The 2D attributes are then lifted to 3D using depth information from a Realsense RGB-D camera. Repeating the process for all candidate object categories results in a 3D spatial-semantic transcript 
of all objects in the scene. \acronym then reasons over this transcript along with the referring expression to identify the object in question. Finally, we compute the pick and place poses accordingly to control the robot's end effector. We provide the pseudo-code below.
\begin{minted}[fontsize=\footnotesize]{python}
def put_first_on_second(self, arg1, arg2):
    # obtain the objects list in the environment
    objs = self.env.get_objs()
    pick_id = get_obj_id(objs, arg1) # Referring
    place_id = get_obj_id(objs, arg2) # Referring
    pick_pose, place_pose = self.get_obj_pose(
        objs, pick_id, place_id)
    self.env.step({'pick': pick_pose, \
        'place': place_pose}) # robot manipulation
\end{minted}

We apply our method to robot manipulation by integrating \acronym as the perception process in the Code-as-Policies (CaP)~\cite{Liang2022CodeAP} framework (``\OurCaP''). As a baseline, we compare to the standard implementation of Code-as-Policies, which uses MDETR~\cite{kamath2021mdetr} for perception (``\VanillaCaP''). When grounding an object in the scene, \VanillaCaP assumes that the object can be uniquely identified by MDETR according to its associated noun phrase (e.g., ``the duckie''). \VanillaCaP then simply selects the output grounding with the highest score. In order to make the comparison fair, we extend CaP's prompt so that it is able to use the entire referring expression as the object identifier, which is indeed unique. Otherwise, CaP would perform poorly in our setting, since it has never seen complicated referring expressions in the original prompts.

\begin{figure}[!t]
     \centering
     \begin{subfigure}[b]{0.49\linewidth}
         \centering
         \includegraphics[width=\textwidth, bb=0 0 244 306]{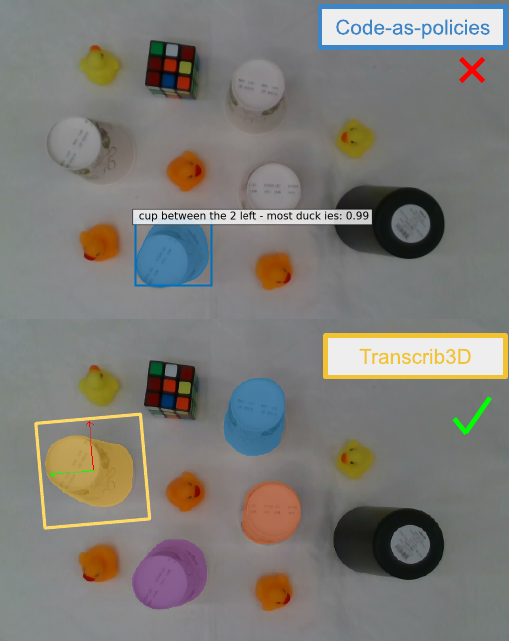}
         \caption{``the cup between the two left-most duckies''}
         \label{fig:cup between the 2 left-most duckies}
     \end{subfigure}
     \begin{subfigure}[b]{0.49\linewidth}
         \centering
         \includegraphics[width=\textwidth, bb=0 0 244 307]{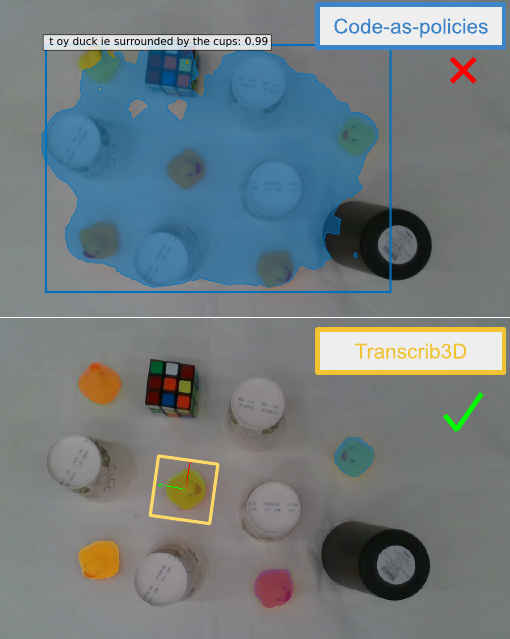}
         \caption{``the toy duckie surrounded by the white cups''}
         \label{fig:toy duckie surrounded by white cups}
     \end{subfigure}
        \caption{Qualitative comparison of the grounding performance of (top) \VanillaCaP and (bottom) \OurCaP on a real robot.}
        \label{fig:real_robot_comparison}
        \vspace{-3ex}
\end{figure}
We evaluate the performance of both methods on five different natural language instructions (each involving two referring expressions). Figure~\ref{fig:real_robot_comparison} shows a qualitative comparison for one of the test cases.
Our results reveal that when assessed at both the referring expression and instruction levels, \OurCaP significantly outperforms \VanillaCaP. \OurCaP achieves an instruction success rate of 80\% (4 out of 5) compared to 20\% (1 out of 5) for \VanillaCaP, and correctly resolves 90\% of the referring expressions (9 out of 10) compared to 40\% (4 out of 10) for \VanillaCaP.
This underscores a pronounced benefit in employing \acronym for resolving referring expressions within robot pick-and-place tasks.

%% file: sections/conclusion.tex
\section{Discussion}

\begin{figure}[!t]
    \centering
    \begin{subfigure}[b]{0.48\linewidth}
         \centering
         \includegraphics[width=\textwidth, bb=0 0 948 587]{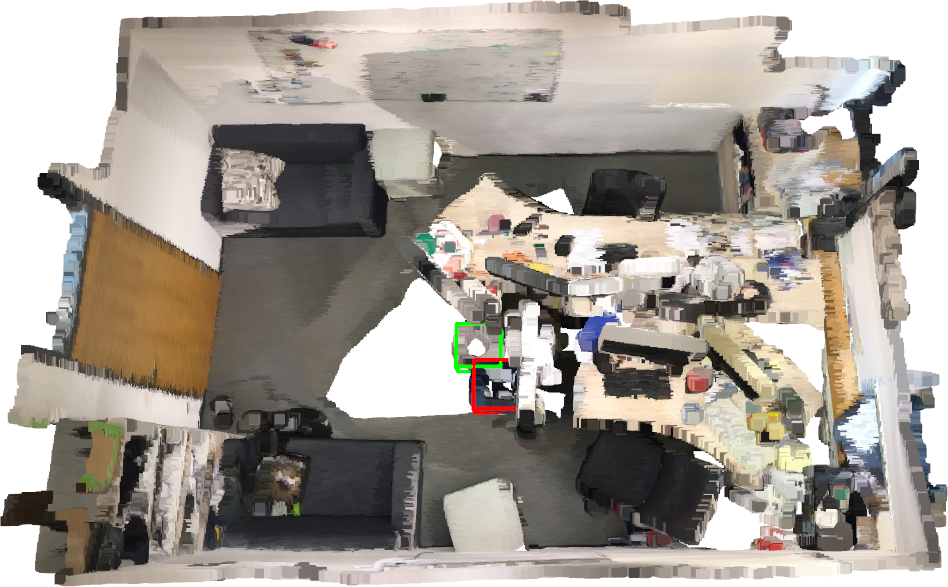}
         \caption{Utterance: ``The cylinder shaped trash can.'' Our model is unable to ground the ``cylinder'' shape, leading to failure.}
         \label{fig:cylinder_failure}
     \end{subfigure}
     \hfill
     \begin{subfigure}[b]{0.48\linewidth}
         \centering
         \includegraphics[width=\linewidth, bb=0 0 1189 687]{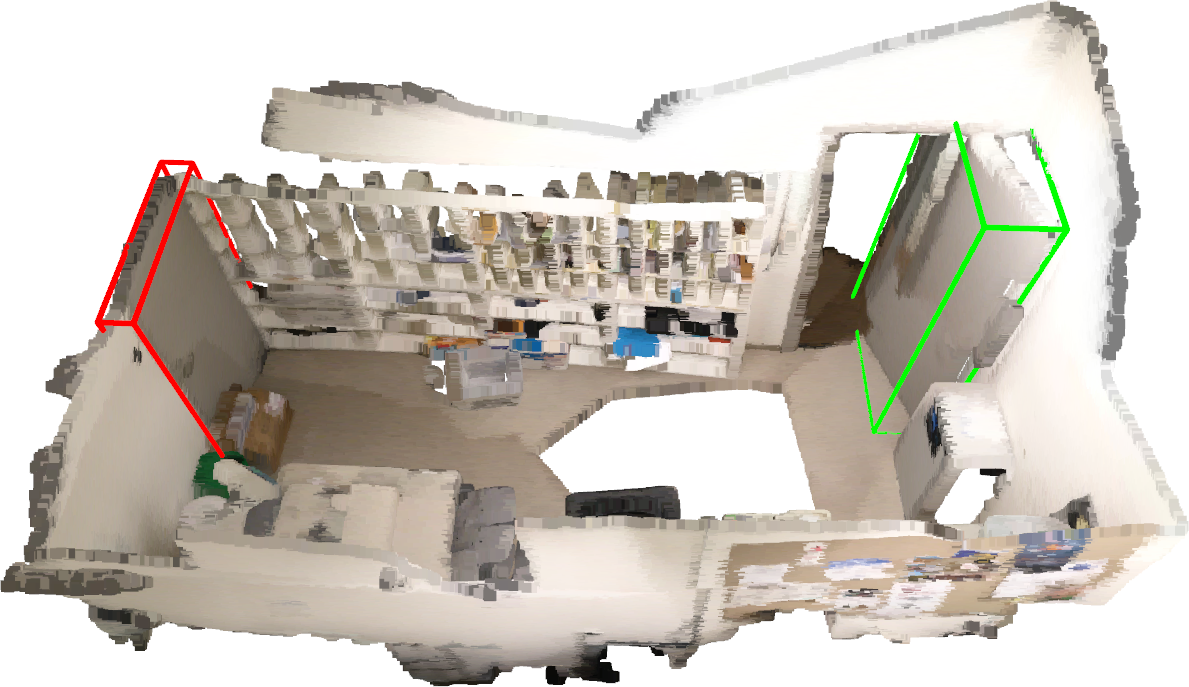}
         \caption{Utterance: ``This door is open and has nothing blocking it.'' This query requires understanding the concept of ``open'', which is hard to infer from a bounding box.}
         \label{fig:failure_2}
     \end{subfigure}
    \caption{Typical failure cases in \acronym, with ground-truth bounding boxes shown in \textcolor{green}{green} and ours in \textcolor{red}{red}.}
    \label{fig:failure_cases}
    \vspace{-3ex}
\end{figure}

We acknowledge that \acronym is not without any limitations.
First, our scene transcript is object-centric. Although object-level details such as bounding boxes are sufficient for numerous 3D spatial reasoning tasks, there exist scenarios that necessitate a finer level of object details. We show examples of two such cases in Figure~\ref{fig:failure_cases}.
Second, our reliance on existing 3D detectors introduces a constraint: the quality of 3D detection itself. In our experiments, we observed that even state-of-the-art 3D detection methods~\cite{Liu2021GroupFree3O, Schult23ICRA} yield sub-optimal results, highlighting the room for improvement in 3D detection. %
Third, our method involves manual specification of the desired information to be extracted from 3D detections (e.g., each object's center, size, and orientation), which has its limitations. An adaptive feature selection strategy could potentially yield better results.

However, even with those limitations, our method surpasses all current multi-modality baselines. This achievement leads us to propose two critical insights: firstly, the connection module facilitating interaction between different modalities may not be as effective since our experiment can be roughly regarded as a controlled test that replaces a typical multi-modal cross-attention module with just \emph{text}, and demonstrates that it works better; and secondly, the reasoning module might be too simplistic to ground the complex logic inherent in natural language.
We believe that the observed limitations in current multi-modality methods can largely be attributed to the scarcity of annotated 3D %
data, which is orders-of-magnitude smaller than its 2D counterpart, primarily due to the higher cost in its collection. We call for more efficient data collection pipelines in 3D and robotics, potentially leveraging generative methods.

\section{Conclusion}

We introduce \acronym, a simple and effective method for the 3D grounding of natural language referring expressions that requires no training, yet delivers state-of-the-art performance across leading 3D reference resolution benchmarks. The idea of using text as a unifying medium to connect the 3D scene and LLM reasoning not only achieves great results, but also provides critical insights into the bottleneck of current multi-modal methods.